\documentclass{article} % For LaTeX2e
\usepackage{iclr2027_conference,times}
\iclrfinalcopy
\usepackage{amsmath,amsfonts,bm}

\def\eqref#1{equation~\ref{#1}}
\def\1{\bm{1}}

\DeclareMathAlphabet{\mathsfit}{\encodingdefault}{\sfdefault}{m}{sl}
\SetMathAlphabet{\mathsfit}{bold}{\encodingdefault}{\sfdefault}{bx}{n}

\usepackage{hyperref}
\usepackage{url}
\usepackage{gensymb}
\usepackage{booktabs}
\usepackage{graphicx}
\usepackage[table,xcdraw]{xcolor}
\usepackage{multirow}
\usepackage[most]{tcolorbox}
\usepackage[most]{tcolorbox}
\usepackage{enumitem}
\usepackage{float}
\usepackage{subcaption}
\usepackage{wrapfig}

\title{Agentic Relative Camera Pose Estimation via Learned Ranking and Verification}

\author{
\textbf{Zhining Gu}$^{1}$ \qquad
\textbf{Shangjie Du}$^{2}$ \qquad
\textbf{Weimin Qiu}$^{2}$ \\
\textbf{Carl Olsson}$^{3}$ \qquad
\textbf{Ping Liu}$^{4}$ \qquad
\textbf{Meng Tang}$^{2}$ \\[0.5em]
$^{1}$Arizona State University \qquad
$^{2}$University of California, Merced \\
$^{3}$Lund University \qquad
$^{4}$University of Nevada, Reno }
\begin{document}

\maketitle
\lhead{}

\begin{abstract}
A wide range of approaches have been developed for camera pose estimation, including correspondence-based methods, end-to-end pose regression, and recent 3D geometric foundation models.
Our key observation is that no single estimator is optimal for diverse challenges, such as wide baselines, lack of texture, appearance changes, and occlusions.
Further analysis reveals substantial performance variation across both benchmarks and individual image pairs, with different estimators exhibiting complementary strengths.
We introduce \textbf{PoseAgent}, an agentic framework for relative camera pose estimation that dynamically orchestrates pose estimators through learnable \textit{ranking} and \textit{verification}.
Given an image pair, a profiling agent first extracts appearance, semantic, and geometric features relevant to pose estimation, e.g., scene type.
A learned ranking agent then predicts the relative competence of multiple pose estimators given the image-pair profile.
The top-ranked estimator is executed, and its predicted pose is assessed by a learned verification agent that estimates the corresponding pose error.
When verification fails, PoseAgent adaptively invokes lower-ranked estimators until a candidate is accepted or the execution budget is reached.
For pose verification, our verification network predicts pose errors more accurately than prior models.
For pose estimation, PoseAgent improves AUC@5\degree up to 4.2\% over the strongest standalone estimator on each of ARKitScenes, MegaDepth, ScanNet++, and RealEstate10K.
On ARKitScenes, PoseAgent also outperforms VLM-based agents, which include a VLM ranker with the same verifier and fallback policy.
These results demonstrate the effectiveness of our learned ranking and verification.
\end{abstract}
%\vspace{-1\baselineskip}

%\vspace{-1\baselineskip}
\section{Introduction}
Camera pose estimation is fundamental to virtual and augmented reality, autonomous navigation, 3D/4D reconstruction, and robotics.
Relative camera pose estimation aims to recover the transformation between the camera coordinate systems of two images.
Existing approaches span sparse keypoint-based methods~\citep{lowe2004distinctive,nister2004efficient,yi2016lift, mishchuk2017working, detone2018superpoint,tian2019sosnet, dusmanu2019d2, sarlin2020superglue, tyszkiewicz2020disk, lindenberger2023lightglue}, dense correspondence methods~\citep{sun2021loftr,edstedt2024roma}, direct pose regression~\citep{kendall2015posenet,dong2025reloc3r}, and recent geometric foundation models~\citep{dust3r_cvpr24,mast3r_eccv24,wang2025vggt}.

Despite substantial progress, relative pose estimation remains challenging under extremely wide baselines, little overlap, repetitive texture, appearance changes, and scene dynamics.
Our systematic evaluation reveals that no single estimator consistently excels across datasets or individual image pairs.
Surprisingly, even classical or seemingly outdated estimators outperform more recent models in many cases, see detailed analysis in Section~\ref{sec:motivation}.
This variation reflects complementary strengths across estimator families.

%
% Different pose estimators exhibit distinct strengths and weaknesses under different challenges.
%
Among these estimator families, correspondence-based methods are highly accurate when reliable feature matches are available, but degrade under large viewpoint changes or limited visual overlap.
In contrast, end-to-end regression methods are more robust for these challenges, while being less accurate than structure-based methods when reliable correspondences are present.
Existing systems partially recognize this model specialization. 
For example, SuperGlue~\citep{sarlin2020superglue} and RoMa~\citep{edstedt2024roma} provide separate models for indoor and outdoor scenes.
However, such dataset-level specialization is too coarse to capture the varying characteristics of individual pairs

\begin{figure}[t!]
    \centering
    \includegraphics[width=0.9\linewidth]{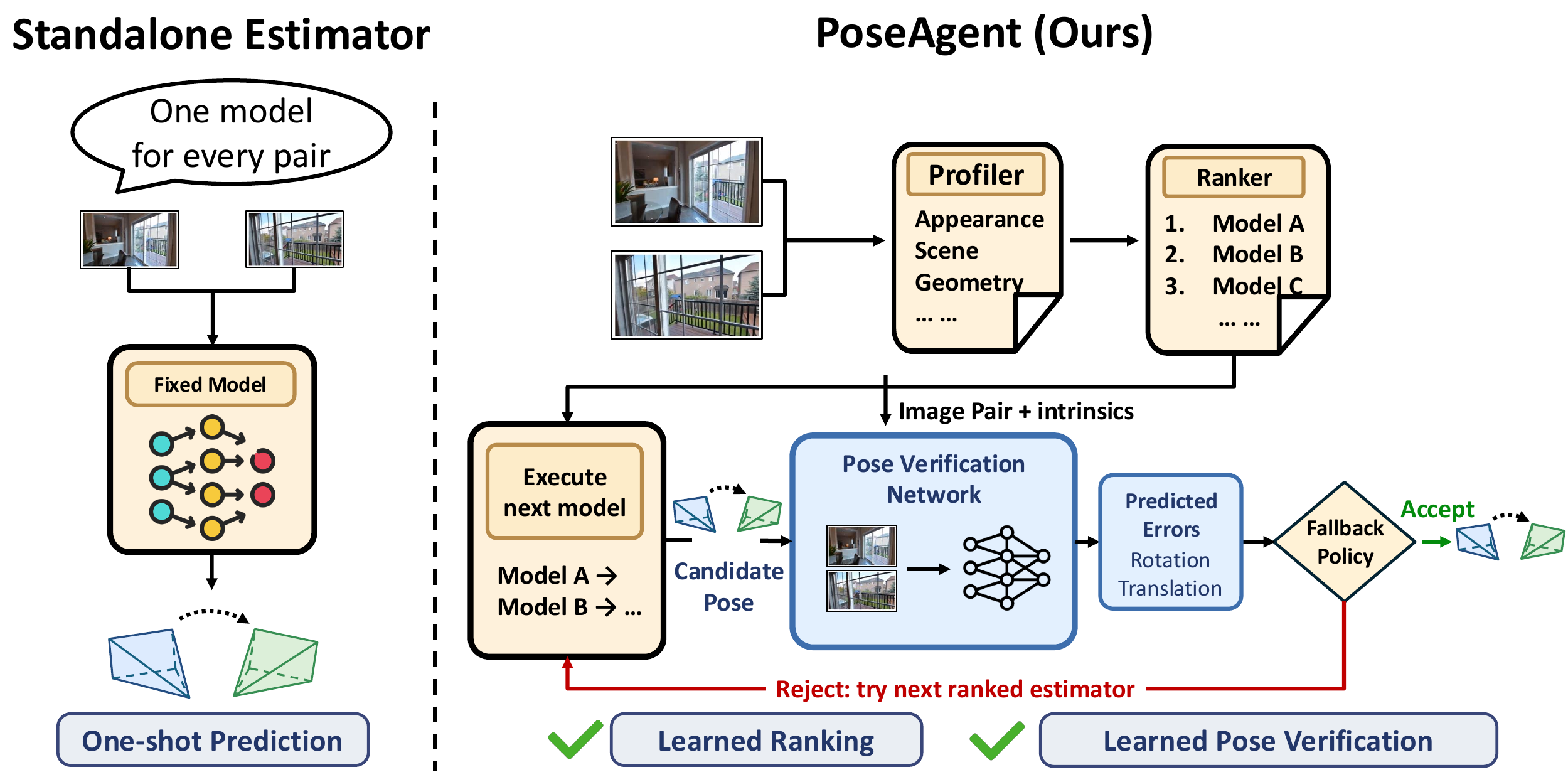}
    \caption{We propose \textbf{PoseAgent}, an agentic framework that adaptively ranks and executes pose estimators and verifies pose estimation with orchestration. Our framework is a meta-learner and improves a set of standalone pose estimators.}
    \label{fig:teaser}
\end{figure}

We therefore boost relative pose estimation at image-pair level via \textbf{instance-adaptive model selection} with \textbf{verification}. 
Given an image pair, a successful system should (1) predict which estimator is likely to be the best given image pair profile and (2) determine whether its output after execution can be trusted without access to ground-truth pose. 
% We therefore ask \textbf{which estimator is most suitable for a particular image pair?}
%
% Answering it requires two capabilities:
%
% (1) predicting which estimator is likely to succeed before execution and
%
% (2) verifying whether its output can be trusted without ground truth pose.
%
%
Towards these goals, we introduce \textbf{PoseAgent}, a closed-loop agentic framework that orchestrates heterogeneous pose estimators as specialized tools, see outline in Fig.~\ref{fig:teaser}.
Inspired by agentic vision systems that coordinate tools through profiling, planning, tool selection, execution, and feedback~\citep{zuo20254kagent, lu2023chameleon}, PoseAgent combines a learned ranker for model selection with a learned verifier for pose verification.

Given an image pair, a profiling agent extracts semantic and geometric characteristics related to visual overlap, appearance variation, scene structure, and matching difficulty. 
Conditioned on this profile, a ranking agent predicts the relative competence of a set of candidate estimators and invokes the highest-ranked one.
Importantly, estimator selection is not treated as a one-shot decision.
Next, the learnable verification agent evaluates the reliability of the predicted pose using geometric evidence, without access to the ground-truth camera pose. 
If the verification fails, PoseAgent adaptively falls back to the next promising estimator according to the ranking. 
In this way, ranking provides a prior estimate of estimator competence, while verification provides feedback about the reliability of the actual prediction, enabling closed-loop recovery from unsuccessful estimator choices.
Our main contributions are summarized as follows:

\begin{itemize}[leftmargin=10pt]
    \item We reveal substantial complementarity among existing pose estimators across datasets and individual image pairs, where even classical estimators outperform recent models in many cases, motivating instance-adaptive model selection.
    \item We introduce \textbf{PoseAgent}, an agentic framework that integrates image-pair profiling, learned estimator ranking, estimator execution, and learned pose verification into a closed-loop inference process with verification-guided fallback to lower-ranked estimators.
    \item Extensive experiments show that \textbf{PoseAgent} outperforms standalone estimators on four datasets and outperforms VLM-based agents on ARKitScenes, including a VLM ranker combined with the same verifier and fallback policy.
\end{itemize}

\section{Related Work}
\label{related_work}

\noindent \textbf{Relative Camera Pose Estimation}
Conventional methods first detect and describe local features using hand-crafted methods, such as SIFT~\citep{lowe2004distinctive}
%FAST~\citep{rosten2006machine}, 
and SURF~\citep{bay2006surf}, establish correspondences through descriptor matching, and estimate relative pose using robust geometric solvers~\citep{multiviewgeometry}.
Deep learning has introduced learned alternatives throughout this pipeline, including detectors and descriptors (e.g., SuperPoint~\citep{detone2018superpoint}), sparse matcher (e.g., SuperGlue~\citep{sarlin2020superglue} and LightGlue~\citep{lindenberger2023lightglue}), and dense matchers (e.g., LoFTR~\citep{sun2021loftr}, and RoMa~\citep{edstedt2024roma}). 
Other geometric primitives beyond keypoint correspondences have also been explored.
Line-based and hybrid point-line approaches improve robustness in weakly textured or structurally dominant scenes~\citep{elqursh2011line, Vakhitov_2018_ECCV, hruby2024handbook}, while more specialized approaches estimate epipolar geometry or relative pose from conics and cylinder silhouettes~\citep{kahl1998conic, Gummeson_2024_ACCV}.
Pose regression methods directly predict relative pose, bypassing explicit correspondence estimation~\citep{kendall2015posenet,arnold2022map,ding2019camnet,khatib2024leveraging,zhou2020learn,winkelbauer2021learning,dong2025reloc3r}.
More recently, geometric foundation models, such as DUSt3R~\citep{dust3r_cvpr24}, Reloc3r~\citep{dong2025reloc3r}, and VGGT~\citep{wang2025vggt} have leveraged Transformer architectures and large-scale 3D data to address many geometric vision tasks. 
These developments have produced a diverse set of estimators with complementary assumptions and failure modes, motivating instance-adaptive model selection.

\noindent \textbf{Model Selection and Pose Verification}
Adaptive selection among pose estimation strategies has been explored in several settings.
\citet{camposeco2018hybrid} perform adaptive solver selection within RANSAC, while \citet{rockwell2024far} learn to balance geometric solver outputs and learned pose predictions. 
\citet{yu2025relative} combine depth-aware and classical point-based solvers through hybrid scoring and refinement, and \citet{panek2026combining} investigate scoring functions for selecting between structure-based and structure-less pose estimates. 
These approaches demonstrate the benefits of model selection, but primarily operate at the solver level or combine a small number of predefined paradigms within a fixed pipeline. 
Complementary to selection, pose verification determines whether a candidate estimate should be trusted.
Classical verification relies on geometric criteria or correspondence consensus, such as Sampson and reprojection errors, or RANSAC consensus scores.
In particular, RANSAC selects hypotheses by inlier count, while \citet{Torr2000MLESAC} replaces this criterion with a likelihood-based objective to reduce sensitivity to hard inlier thresholds.
However, such criteria depend on the quality of the underlying correspondences. 
Learning-based methods instead predict hypothesis quality from visual and geometric evidence. 
For example, FSNet~\citep{barroso2023two} scores fundamental matrix hypotheses from an image pair without relying on sparse correspondence features.
In contrast to prior work, we rank multiple estimators spanning correspondence-based methods, direct pose regression, and large-scale geometric models, and verify their final pose predictions within a closed-loop fallback process. 
%
%The ranker and verifier thus address two complementary questions: which estimator is most suitable before execution, and whether its output should be accepted afterward.

\noindent \textbf{Multimodal Large Language Model (MLLM) Agent}
LLMs have enabled agents capable of planning, reasoning, and tool use~\citep{yao2022react, hong2023metagpt, wang2023describe, hu2025g2vlmgeometrygroundedvision, zuo20254kagent, szot2025multimodal, xu2026area3d, yao2026photoagent, zhang2026predicting}.
Multimodal agents have evolved from mediating visual inputs through textual descriptions or external perception modules~\citep{wu2023visual,suris2023vipergpt, yang2023mm, gao2023assistgpt} to jointly processing visual and textual inputs with native multimodal models.
However, general-purpose multimodal representations often lack the precise geometric and spatial information required by vision tasks~\citep{deng2026lostspacevisionlanguagemodels, ma2024spatialpin, guo2025pursuing, marsili2025visual}.
Tool-augmented agents address this limitation by leveraging external vision or geometry models during reasoning~\citep{cho2026spatialclaw, guo2025pursuing, zuo20254kagent, wang2025vggt, kirillov2023segment, guo2025pursuing, ma2024spatialpin, marsili2025visual}. 
Our work follows this tool-augmented agentic paradigm for pose estimation by including heterogeneous pose estimators as specialized tools.
Unlike general-purpose multimodal LLM, PoseAgent specializes the orchestration process for relative camera pose estimation through learned ranking and verification.
\section{Method}
\label{method}

\subsection{Motivating Analysis}
\label{sec:motivation}

\begin{figure}[t!]
    \centering
    \includegraphics[width=\linewidth]{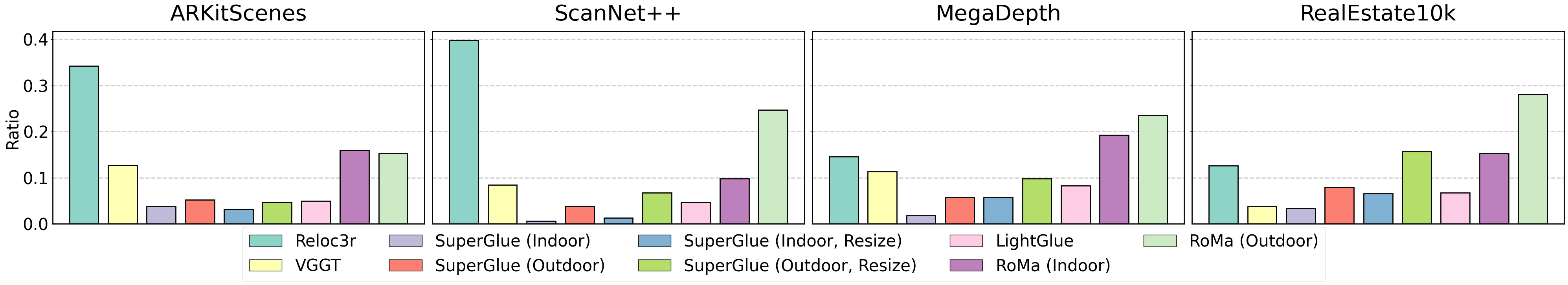}
    \caption{\textbf{Distributions of best estimator across various datasets.} No single estimator dominates the lowest pose error across all datasets or image pairs.}
    % \caption{\textbf{Frequency of each estimator produces the lowest pose error on each dataset.} Although some estimators are selected more frequently on each dataset, no single estimator consistently performs the best across all image pairs.}
    % \caption{The histogram records the frequency of each estimator when it gives the minimum pose error on an image pair. Seven methods are benchmarked on three evaluation datasets.}
    
    \label{fig:best freq}
\end{figure}

Recent estimators have achieved strong benchmark performance~\citep{lindenberger2023lightglue,edstedt2024roma,dong2025reloc3r,wang2025vggt}, yet no single estimator is the most accurate across image pairs.
To examine this, Figure~\ref{fig:best freq} reports, for each dataset, the fraction of image pairs on which each estimator is the most accurate, i.e., produces the lowest pose error.
This fraction is highest for Reloc3r on the indoor ARKitScenes~\citep{baruch2021arkitscenes} and ScanNet++~\citep{yeshwanth2023scannet++} datasets, and for RoMa Outdoor~\citep{edstedt2024roma} on MegaDepth~\citep{li2018megadepth} and RealEstate10K~\citep{zhou2018stereo}.
However, even these most frequent winners are the most accurate estimator on fewer than 40\% of the image pairs in any dataset, and on only about 23\% of the pairs on MegaDepth.
The remaining pairs are often won by older estimators: the four SuperGlue variants together are the most accurate on about one third of the RealEstate10K pairs.

This per-pair variation translates into a large accuracy gap: an oracle that selects the most accurate of the nine estimators for each pair outperforms the strongest standalone estimator by 10.5 to 15.1 percentage points in AUC@5\degree across the four datasets (Table~\ref{tab:eval_ranker_model_selection}).
Such per-pair selection cannot be reduced to choosing by scene type, since Reloc3r and RoMa Outdoor each win a substantial fraction of pairs on both indoor and outdoor datasets.
Estimator suitability therefore likely depends on pair-specific factors beyond scene type, such as viewpoint change, visual overlap, texture, and appearance variation, which is consistent with our profiling ablation in Section~\ref{sub:ablation}.
Closing the oracle gap without ground truth requires two capabilities: predicting a suitable estimator for each image pair from these factors, and verifying its output.

\subsection{Problem Formulation}
\label{sec:formulation}

Both capabilities operate on the outputs of individual pose estimators.
Given an image pair $(I_1, I_2)$ and the corresponding camera intrinsics $(\mathbf{K}_1,\mathbf{K}_2)$, a pose estimator $M_m$ predicts the relative camera pose between the two camera coordinate systems:
\begin{equation}
    M_m(I_1,I_2;\mathbf{K}_1,\mathbf{K}_2)
    \longrightarrow
    \widehat{\mathbf{T}}_{21}^{m} = [\widehat{\mathbf{R}}_{21}^{m}\mid\widehat{\mathbf{t}}_{21}^{m}]
\end{equation}
where $\widehat{\mathbf{R}}_{21}^{m}$ denotes the rotation matrix and $\widehat{\mathbf{t}}_{21}^{m}$ denotes the translation vector.
PoseAgent is a meta-method that treats an arbitrary set of such estimators, $\mathcal{M}=\{M_1,\ldots,M_N\}$, as callable tools and adaptively coordinates them:
\begin{equation}
    [\widehat{\mathbf{R}}_{21}^{*}\mid\widehat{\mathbf{t}}_{21}^{*}]
    =
    \mathcal{A}(I_1,I_2,\mathbf{K}_1,\mathbf{K}_2;\mathcal{M},K)
\end{equation}
where $K\leq N$ is the maximum number of estimator executions.
The ranked execution route and verification feedback jointly determine the final pose estimate.

\begin{figure}[t!]
    \centering % Centers the image
    \includegraphics[width=0.9\textwidth]{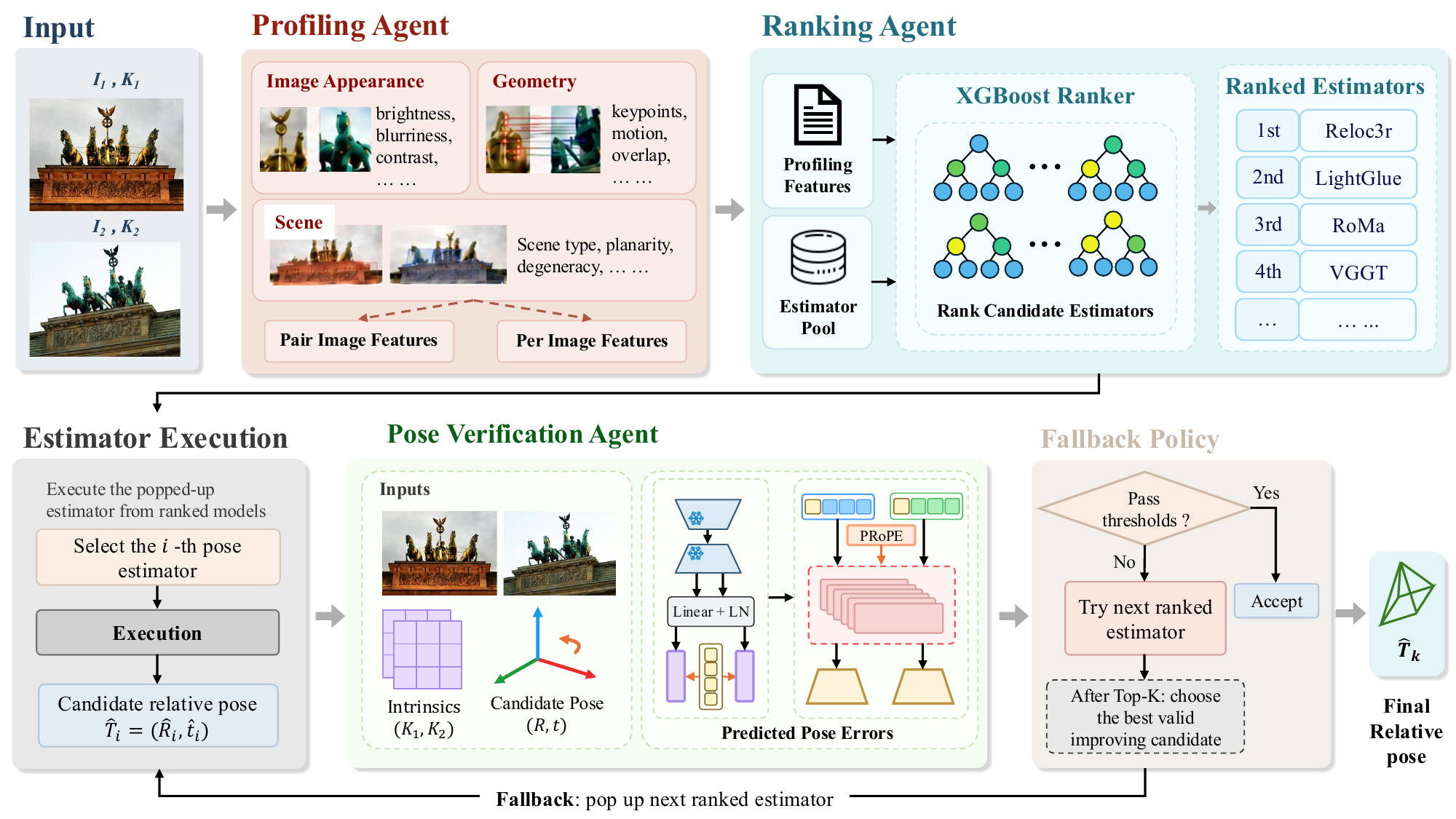} % File name without extension
    \caption{\textbf{PoseAgent} profiles an image pair, ranks and sequentially executes candidate estimators, and verifies each prediction until a pose is accepted following a fallback policy.}
    \label{fig:architecture} % Used to reference the image in the text
\end{figure}

Figure~\ref{fig:architecture} shows how PoseAgent, as a closed-loop agentic framework, produces this route and feedback for each image pair.
The \textbf{Image Pair Profiling Agent} (Sec.~\ref{sec:profiling}) extracts appearance, semantic, and geometric features, which the \textbf{Ranking Agent} (Sec.~\ref{sec:ranking}) uses to rank candidate estimators into an execution route.
For each executed estimator, the \textbf{Pose Verification Agent} (Sec.~\ref{sec:verification}) predicts the rotation and translation errors of its pose.
Finally, the verification-guided fallback policy (Sec.~\ref{sec:fallback}) uses these predicted errors to decide whether to accept the current pose or invoke the next estimator, up to the execution budget $K$.

%
% Fig.~\ref{fig:architecture} shows an overview of our framework.

\subsection{Image Pair Profiling Agent}
\label{sec:profiling}

The {Profiling Agent} constructs a structured feature profile for each image pair using OpenCV, CLIP~\citep{radford2021learning}, DINOv2~\citep{oquab2023dinov2}, SegFormer~\citep{xie2021segformer}, LoFTR~\citep{sun2021loftr}, and SuperPoint~\citep{detone2018superpoint}. 
The extracted appearance, semantic, and geometric features characterize individual images and their cross-view relationships, including scene composition, appearance changes, and matching conditions.
The resulting profile is passed to the Ranking Agent to rank the candidate estimators. 
Appendix~\ref{app:profiling_tools} provides the complete list of tools and features. 

% The {Profiling Agent} constructs a structured description of each image pair to support model selection. 
%
% To extract appearance statistics, semantic information, and correspondence-based geometric cues, it executes a collection of visual analysis tools, including OpenCV tools, CLIP~\citep{radford2021learning}, DINOv2~\citep{oquab2023dinov2}, SegFormer~\citep{xie2021segformer}, LoFTR~\citep{sun2021loftr}, and SuperPoint~\citep{detone2018superpoint}.
%
% These features characterize both individual images and their cross-view relationships, including scene composition, appearance changes, and matching conditions.
%
% The list of computer vision tools we use is provided in the Appendix~\ref{app:profiling_tools}.
%
% The resulting profile is further passed to the {Ranking Agent}, which uses it to rank all estimators.

\subsection{Ranking Agent}
\label{sec:ranking}

Given the image-pair features produced by the {Profiling Agent}, {Ranking Agent} estimates the suitability of each pose estimator and generates a ranking.
%
%We formulate estimator selection as a label ranking problem, since PoseAgent requires an ordered list of candidate estimators for subsequent verification and fallback. 
Specifically, given image pair profile $\mathbf{p}_i$ augmented with candidate estimator identity for image pair $i$, we train a light-weight XGBRanker~\citep{chen2016xgboost} $r_\theta$, which is a tree-based ranking model, to predict an estimator suitability score $s_{i}^{m}=r_\theta\left(\mathbf{p}_{i}^{m}\right)$ for a model $M_m$.
%where $m$ identifies the candidate estimator. 
% \begin{equation}
%     s_{i}^{m}=r_\theta\left(\mathbf{p}_{i}^{m}\right)
% \end{equation}
% We then sort the candidate estimators in descending order of their predicted suitability scores and plan the execution order.
Sorting the candidates by decreasing $s_i^m$ produces the execution order, prioritizing suitable estimators without exhaustively executing all candidates.
% \begin{equation}
%     \mathcal{R}_i
%     = 
%     \left(
%         M_i^{[1]}, M_i^{[2]}, \ldots, M_i^{[N]}
%     \right)
% \end{equation}
% where $M_i^{[k]}$ denotes the estimator ordered to the $k$-th position for image pair $i$, such that $s_i^{[1]} \geq s_i^{[2]} \geq \cdots \geq s_i^{[N]}$.

% Overall, the {Ranking Agent} plans the tool execution order for each image pair, which can further be used sequentially by {Verification Agent} and fallback.
%
% Given the structured profiling features produced by the \textbf{Profiling Agent}, the \textbf{Ranking Agent} estimates the suitability of each pose estimator and produces a prioritized execution routes.
% We formulate this process as a label ranking problem, since PoseAgent requires an ordered list of candidate estimators for subsequent verification and fallback. 

% Instead of exhausting all candidate pose estimators, {Ranking Agent} is expected to place most suitable models in the first place, which avoids expensive model execution and computation. 

\noindent \textbf{Training data for ranking} For each image pair $i$ along with candidate estimator $M_m$, we define its pose error as $e_{i}^{m} = 
    \max\left(e_{R, i}^{m}, e_{t, i}^{m}\right)$, 
% \begin{equation}
%     e_{i}^{m} = 
%     \max\left(e_{R, i}^{m}, e_{t, i}^{m}\right)
% \end{equation}
where $e_{R, i}^{m}$ and $e_{t, i}^{m}$ denote the rotation and translation direction errors with respect to the ground truth relative pose. We rank the $N$ candidate estimators in ascending order of $e_{i}^{m}$ and convert their ranks into relevance labels, assigning higher relevance to lower error candidates, to train an XGBRanker.
% \begin{equation}
%     y_{i}^{m}=N+1-r_{i}^{m}
% \end{equation}
% where $r_{i}^{m}$ is the rank of estimator $M_m$ based on pose errors.
%Estimators with lower pose errors receive higher relevance labels. 

% For each candidate, we combine the image profiling features $\mathbf{p}_i$ with an one-hot encoding of the estimator:
% \begin{equation}
%     \mathbf{x}_{i}^{m}
%     =
%     \left[
%         \mathbf{p}_i;
%         \operatorname{one-hot}(M_i^{[m]})
%     \right].
% \end{equation}

\subsection{Pose Verification Agent}
\label{sec:verification}

Our Verification Agent evaluates each pose hypothesis: after each estimator along the ranked route is executed, it predicts the errors of the candidate pose.
For image pair $i$, let the estimator at the $k$-th position of the execution route produce a candidate relative pose $\widehat{\mathbf{T}}_{i}^{[k]}=[\widehat{\mathbf{R}}_{i}^{[k]}\mid\widehat{\mathbf{t}}_{i}^{[k]}]$.
We train a \textbf{Pose Verification Network (PVN)} $V_{\phi}$ to predict its rotation and translation direction errors:
\begin{equation}
\left(
\widehat{e}_{R,i}^{[k]},
\widehat{e}_{t,i}^{[k]}
\right)
=
V_{\phi}
\left(
I_{1,i}, I_{2,i},
\mathbf{K}_{1,i}, \mathbf{K}_{2,i},
\widehat{\mathbf{T}}_{i}^{[k]}
\right)
\end{equation}

\begin{figure}[t!]
    \centering
    \includegraphics[width=0.85\linewidth]{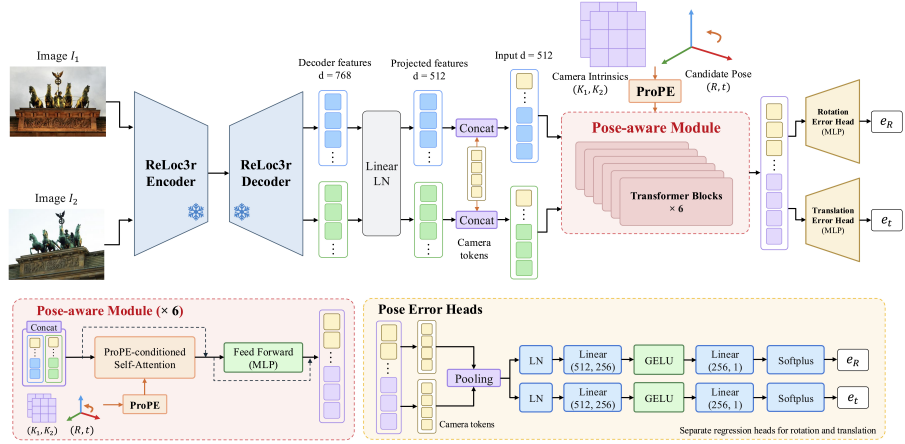}
    \caption{\textbf{Architecture of Pose Verification Network (PVN).} The network is conditioned on camera intrinsics and a candidate pose to predict rotation and translation errors.}
    \label{fig:verifier}
\end{figure}

\paragraph{Network architecture.}
As shown in Fig.~\ref{fig:verifier}, PVN builds upon the pretrained Reloc3r encoder and decoder~\citep{dong2025reloc3r}, whose parameters remain frozen during training.
% \todo{One sentence on why Reloc3r is used as the backbone, and a note that it is also one of the candidate estimators.}
We use Reloc3r as the backbone because it is pretrained for pose regression, so its decoder features already encode cross-view geometry without relying on explicit correspondences.
Although it is also one of the nine candidate estimators, PVN only reuses its image pair features, while the verification decision is conditioned on each candidate pose and applied to all estimators in the same way. 
On top of the frozen decoder, we append six trainable Transformer blocks~\citep{kang2025multi} that alternate between within-view attention and pose-conditioned cross-view attention, in which the candidate pose and camera intrinsics are embedded using \textbf{PRoPE}~\citep{li2025cameras}.
Learnable register tokens~\citep{kang2025multi} prepended to the visual tokens of each view aggregate these pose-conditioned features.
The pooled register tokens of the two views are fed to two regression heads with Softplus activations, which predict non-negative rotation and translation direction errors.% the translation head additionally takes $\widehat{\mathbf{t}}_{i}^{[k]}$ as input.

\noindent\textbf{Training objective.}
PVN is supervised with the ground-truth rotation and translation direction errors of each candidate pose.
To reduce the influence of the long-tailed error distribution and emphasize accuracy on small errors, we minimize an $\ell_1$ loss in logarithmic space:
\begin{equation}
\mathcal{L}_{\mathrm{PVN}}
=
\left|
\log(1+\widehat{e}_{R})-\log(1+e_R)
\right|
+
\lambda_t
\left|
\log(1+\widehat{e}_{t})-\log(1+e_t)
\right|
\end{equation} 
% \todo{confirm whether or not  l1 is used here}
where $\lambda_t$ controls the relative contribution of translation error regression.
At inference, the predicted errors of each candidate are passed to the fallback policy.

\subsection{Closed-Loop Inference Pipeline}
\label{sec:fallback}

The fallback policy connects the three agents into a closed loop at inference time.
For image pair $i$, the Ranking Agent orders the candidate estimators using the profile $\mathbf{p}_i$, and PoseAgent executes at most $K \leq N$ of them along this route, passing each candidate to the Verification Agent and each prediction to the fallback policy.
For the candidate $\widehat{\mathbf{T}}_i^{[k]}$ at step $k$, the policy first checks whether its predicted errors satisfy the verification thresholds $\tau_R$ and $\tau_t$:
\begin{equation}
\label{eq:fallback}
q_i^{[k]}
=
\mathbb{I}
\left[
\widehat{e}_{R,i}^{[k]} < \tau_R
\;\land\;
\widehat{e}_{t,i}^{[k]} < \tau_t
\right]
\end{equation}
For $k>1$, it additionally checks whether the candidate improves upon the Rank-1 hypothesis in both error dimensions:
\begin{equation}
b_i^{[k]}
=
\mathbb{I}
\left[
\widehat{e}_{R,i}^{[k]} <
\widehat{e}_{R,i}^{[1]}
\;\land\;
\widehat{e}_{t,i}^{[k]} <
\widehat{e}_{t,i}^{[1]}
\right],
\qquad k>1.
\end{equation}
PoseAgent accepts the Rank-1 candidate if $q_i^{[1]}=1$ and a subsequent candidate if $q_i^{[k]}b_i^{[k]}=1$; otherwise, it executes the next estimator along the route until the budget $K$ is reached.
If no candidate is accepted, PoseAgent returns the candidate with the smallest predicted error $\max(\widehat{e}_{R,i}^{[k]},\widehat{e}_{t,i}^{[k]})$ among those with $b_i^{[k]}=1$, or the Rank-1 candidate if no such candidate exists.
This closed loop makes PoseAgent agentic in the sense that it does not apply a fixed estimator or exhaustively run all models, but adapts its sequence of estimator calls to the verification feedback on each image pair.

\section{Experiments}
\label{experiments}

% \subsection{Experimental Setup}
% \label{sub:setup}
\noindent\textbf{Datasets.}
We evaluate PoseAgent on four datasets covering indoor and outdoor scenes: ARKitScenes~\citep{baruch2021arkitscenes}, ScanNet++~\citep{yeshwanth2023scannet++}, MegaDepth~\citep{li2018megadepth}, and RealEstate10K~\citep{zhou2018stereo}.
We follow the split used by Reloc3r~\citep{dong2025reloc3r} for ARKitScenes and the provided split for RealEstate10K, and construct scene-disjoint training and test splits for MegaDepth and ScanNet++.
For MegaDepth and ScanNet++, both the training and evaluation splits are drawn from scenes included in Reloc3r pretraining, while ARKitScenes and RealEstate10K use the held-out evaluation splits adopted by Reloc3r.
Since Reloc3r is used as a frozen backbone in PVN, we additionally evaluate verification-guided fallback on the standard ScanNet1500 and MegaDepth1500 benchmarks~\citep{dong2025reloc3r} and assess its generalization beyond these splits.
The number of image pairs in each split is reported in Appendix~\ref{app:implementation_details}.

\noindent\textbf{Evaluation Metrics.}
We report the area under the cumulative error curve (AUC) at 5\degree, 10\degree, and 20\degree (AUC@5\degree, AUC@10\degree, and AUC@20\degree), where pose error is the maximum of the rotation and translation errors~\citep{sarlin2020superglue, dong2025reloc3r}.
For pose verification, we additionally report mAA@10\degree and median errors for rotation, translation, and their maximum~\citep{barroso2023two, dong2025reloc3r, jin2021image}.

\noindent\textbf{Candidate Pose Estimators.}
PoseAgent operates over nine candidate estimators, including four SuperGlue variants using indoor or outdoor weights, with or without image resizing (SG-I, SG-O, SG-I-RS, and SG-O-RS)~\citep{sarlin2020superglue}, LightGlue~\citep{lindenberger2023lightglue}, two RoMa variants (RoMa-I and RoMa-O)~\citep{edstedt2024roma}, Reloc3r~\citep{dong2025reloc3r}, and VGGT~\citep{wang2025vggt}.
I/O denotes indoor/outdoor weights, and RS denotes resizing the longer image side.

\subsection{Main Results}
\label{sub:main_results}
We compare PoseAgent with nine standalone estimators, each of which applies a fixed model to every image pair.
As shown in Table~\ref{tab:main_results}, the strongest standalone estimator differs across datasets: Reloc3r is the best on ARKitScenes and ScanNet++, whereas RoMa-O is the best on MegaDepth and RealEstate10K.
PoseAgent outperforms the best standalone estimator on all four datasets and at all AUC thresholds, without knowing in advance which estimator is the best for a given dataset.
The AUC@5\degree improvement ranges from 0.9\% on MegaDepth to 4.2\% on ScanNet++.
These results demonstrate that instance-adaptive selection and verification can exploit estimator complementarity more effectively than relying on any single fixed estimator.
More qualitative examples are shown in Appendix~\ref{app:visualization}.
%define colors
\definecolor{rankerbg}{RGB}{255, 226, 153} % light yellow 
\definecolor{verifierbg}{RGB}{252, 229, 205} % light pink 
\definecolor{numred}{RGB}{255, 0, 0}
\definecolor{numgreen}{RGB}{112, 173, 71}
\definecolor{numblue}{RGB}{0, 0, 255}
\definecolor{lightgray}{gray}{0.9}

\begin{table*}[t]
    \centering
    % \caption{\textbf{Relative camera pose estimation across four datasets.} The best results for each individual method ar highlighted in \textcolor{numblue}{blue}. Our results are in \textcolor{numred}{red} if they are outperform above results.}
    % \caption{\textbf{Relative camera pose estimation across four datasets.} We report AUC at error thresholds of $5\degree$/$10\degree$/$20\degree$. Best overall results are shown in \textbf{bold}, and the best standalone baseline is \underline{underlined}.}
    \caption{\textbf{Relative pose AUC on four datasets at 5\degree/10\degree/20\degree}. \textbf{Bold} indicates the best result, and \underline{underline} indicates the best standalone estimator.}
    \label{tab:main_results}
    \resizebox{\textwidth}{!}{
    \begin{tabular}{lcccccccccccc}
        \toprule

        & \multicolumn{3}{c}{\textbf{ARKitScenes}}
        & \multicolumn{3}{c}{\textbf{MegaDepth}}
        & \multicolumn{3}{c}{\textbf{ScanNet++}}
        & \multicolumn{3}{c}{\textbf{RealEstate10K}} \\

        \cmidrule(lr){2-4}
        \cmidrule(lr){5-7}
        \cmidrule(lr){8-10}
        \cmidrule(lr){11-13}

        \textbf{Method}
        & \textbf{@5\degree} & \textbf{@10\degree} & \textbf{@20\degree}
        & \textbf{@5\degree} & \textbf{@10\degree} & \textbf{@20\degree}
        & \textbf{@5\degree} & \textbf{@10\degree} & \textbf{@20\degree}
        & \textbf{@5\degree} & \textbf{@10\degree} & \textbf{@20\degree} \\
        \midrule

        Reloc3r
        % & \textcolor{numblue}{\bf 0.456} & \textcolor{numblue}{\bf 0.675} & \textcolor{numblue}{\bf 0.819}
        & \underline{0.456} & \underline{0.675} & \underline{0.819} 
        & 0.562 & 0.732 & 0.847 
        & \underline{0.684} & \underline{0.833} & \underline{0.915}
        & 0.605 & 0.764 & 0.861 \\

        VGGT
        & 0.243 & 0.455 & 0.647 
        & 0.530	& 0.687	& 0.804	
        & 0.194	& 0.342	& 0.515	
        & 0.323	& 0.539	& 0.712\\

        SG-I
        & 0.161	& 0.296	& 0.451	
        & 0.162	& 0.257	& 0.355	
        & 0.094	& 0.184	& 0.299	
        & 0.359	& 0.524	& 0.662 \\

        SG-O
        & 0.242	& 0.394	& 0.539	
        & 0.430	& 0.572	& 0.686	
        & 0.258	& 0.378	& 0.485	
        & 0.568	& 0.706	& 0.802 \\

        SG-I-RS
        & 0.146	& 0.264	& 0.393	
        & 0.320	& 0.413	& 0.496	
        & 0.152	& 0.227	& 0.308	
        & 0.467	& 0.596	& 0.696 \\

        SG-O-RS
        & 0.208	& 0.345	& 0.474	
        & 0.542	& 0.668	& 0.764	
        & 0.302	& 0.407	& 0.498	
        & 0.631	& 0.749	& 0.828 \\

        LightGlue
        & 0.228	& 0.389	& 0.535	
        & 0.511	& 0.647	& 0.745	
        & 0.289	& 0.400	& 0.495	
        & 0.520	& 0.669	& 0.774 \\

        RoMa-I
        & 0.378	& 0.561	& 0.702	
        & 0.662	& 0.776	& 0.854	
        & 0.440	& 0.582	& 0.699	
        & 0.649	& 0.778	& 0.861 \\

        RoMa-O
        & 0.372	& 0.543	& 0.678	
        % & \textcolor{numblue}{\bf 0.684} & \textcolor{numblue}{\bf 0.793} & \textcolor{numblue}{\bf 0.867}
        & \underline{0.684} & \underline{0.793} & \underline{0.867}
        & 0.547	& 0.672	& 0.762	
        % & \textcolor{numblue}{\bf 0.724} & \textcolor{numblue}{\bf 0.827} & \textcolor{numblue}{\bf 0.891} \\
        & \underline{0.724} & \underline{0.827} & \underline{0.891} \\

        % \rowcolor{rankerbg}
        \midrule
        % \textbf{Ranker} (Ours)
        % & \textcolor{numgreen}{0.460}	& 0.674	& 0.816	
        % & 0.682	& \textcolor{numgreen}{0.794}	& \textcolor{numgreen}{0.869}
        % & \textcolor{numgreen}{0.702}	& \textcolor{numgreen}{0.839}	& \textcolor{numgreen}{0.916}
        % & \textcolor{numgreen}{0.729}	& \textcolor{numgreen}{0.836}	& \textcolor{numgreen}{0.901} \\

        \rowcolor{lightgray}
        \textbf{PoseAgent} (Ours)
        & {\bf 0.473} & {\bf 0.685}	& {\bf 0.823}
        & {\bf 0.693} & {\bf 0.810}	& {\bf 0.889}
        & {\bf 0.726} & {\bf 0.855} & {\bf 0.926}
        & {\bf 0.736} & {\bf 0.843}	& {\bf 0.907} \\

        \bottomrule
    \end{tabular}
    
    }
\end{table*}

\subsection{Component Analysis}
\label{sub:component_analysis}

\subsubsection{Ranking Agent}
\label{sub:eval_ranker}

% \noindent\textbf{Ranking Agent.}
We evaluate whether the Ranking Agent prioritizes suitable estimators among the nine candidates.
For $k \in \{1,\ldots,9\}$, Rank-$k$ denotes the prediction produced by the estimator at predicted rank $k$, and Oracle@$k$ selects the best pose among the top-$k$ ranked estimators using the ground-truth pose, serving as an upper bound for selection within the top $k$.
As shown in Table~\ref{tab:eval_ranker_model_selection}, Rank-1 is competitive with or better than the strongest standalone estimator in Table~\ref{tab:main_results}.
The improvement is most evident on ScanNet++ and RealEstate10K, while performance on ARKitScenes and MegaDepth remains comparable to the strongest standalone baselines.
% merge table 3 and table 6 together 
\begin{table}[t]
    \centering
    \caption{\textbf{Ranking and model selection.} Results are AUC@5\degree/10\degree/20\degree. Verifier-Only runs all 9 estimators and picks the lowest predicted error. Fixed-Order + PVN uses one estimator order for all datasets (average win rate from train set) with the same PVN, fallback policy, and $K=4$.}
    \label{tab:eval_ranker_model_selection}
    \scriptsize
    \setlength{\tabcolsep}{4pt}
    \renewcommand{\arraystretch}{0.95}
    \begin{tabular}{lcccc}
        \toprule
        \textbf{Method}
        & \textbf{ARKitScenes}
        & \textbf{MegaDepth}
        & \textbf{ScanNet++}
        & \textbf{RealEstate10K} \\
        \midrule
        Rank-1 (Ranker-Only)
        & {0.460 / 0.674 / 0.816}
        & {0.682 / 0.794 / 0.869}
        & {0.702 / 0.839 / 0.916}
        & {0.729 / 0.836 / 0.901} \\

        Rank-2
        & 0.393 / 0.591 / 0.742
        & 0.655 / 0.774 / 0.855
        & 0.543 / 0.680 / 0.779
        & 0.682 / 0.805 / 0.882 \\
        \midrule

        Verifier-Only
        & 0.455 / 0.669 / 0.813
        & 0.668 / 0.794 / 0.879
        & \textbf{0.728	/ 0.856 / 0.927}
        & 0.698	/ 0.820	/ 0.894 \\

        Fixed-order Ranking + PVN
        & 0.467	/ 0.681	/ 0.821
        & 0.611	/ 0.762	/ 0.863
        & 0.694	/ 0.838	/ 0.917
        & 0.636	/ 0.786	/ 0.875 \\ 

        % \rowcolor{gray!10}
        \rowcolor{red!10}
        \textbf{PoseAgent (Our Ranker + PVN)}
        & \textbf{0.473} / \textbf{0.685} / \textbf{0.823}
        & \textbf{0.693} / \textbf{0.810} / \textbf{0.889}
        & 0.726	/ 0.855	/ 0.926	
        & \textbf{0.736} / \textbf{0.843} / \textbf{0.907} \\

        \midrule
        \rowcolor{gray!10}
        Oracle@2
        & 0.532 / 0.727 / 0.851 
        & 0.715 / 0.819 / 0.888 
        & 0.762 / 0.876 / 0.937 
        & 0.774 / 0.867 / 0.921 \\
        
        \rowcolor{gray!7}
        \textit{Oracle@9}
        & 0.607	/ 0.778	/ 0.881	
        & 0.803	/ 0.888	/ 0.940	
        & 0.800	/ 0.897	/ 0.948	
        & 0.829	/ 0.900	/ 0.941 \\

        \bottomrule
    \end{tabular}
\end{table}
Rank-1 also consistently outperforms Rank-2 across all datasets, indicating that the ranker places stronger estimators earlier.
The gap between Rank-1 and Oracle@2 shows that the second-ranked estimator is more accurate than the first on a subset of pairs, which motivates verification-guided fallback.

\subsubsection{Pose Verification Agent}
\label{sub:eval_verifier}

% \noindent\textbf{Pose Verification Agent}
\noindent\textbf{Error Prediction and Pose Selection.}
% \paragraph{Error Prediction and Pose Selection.}
Table~\ref{tab:verifier_mAA_mae} compares PVN with FSNet~\citep{barroso2023two} on candidate poses for which both methods return valid predictions.
PVN achieves lower MAE for rotation, translation direction, and their maximum.
When selecting among all nine candidates by the smallest predicted maximum error, PVN also improves all mAA and median error metrics over FSNet.

\noindent\textbf{Verification on Standard Benchmarks.}
We further evaluate verification-guided fallback on the ScanNet1500~\citep{sarlin2020superglue} and MegaDepth1500~\citep{sun2021loftr} benchmarks.
For each benchmark, we keep the initial estimator fixed and randomly sample two orders for the remaining candidates, using the same PVN, thresholds ($\tau_R=0.5^\circ$ and $\tau_t=6^\circ$), fallback policy, and $K=4$.
As shown in Table~\ref{tab:verifier_on_1500}, all PVN-based variants outperform the initial estimator alone across both benchmarks and all AUC thresholds, showing that verification remains effective across different execution orders.

\begin{table}[t]
    \centering
    \caption{\textbf{Pose error prediction and selection on MegaDepth and ScanNet++.}}
    \label{tab:verifier_mAA_mae}
    \scriptsize
    \setlength{\tabcolsep}{4pt}
    \renewcommand{\arraystretch}{0.95}

    \begin{tabular}{llccc}
        \toprule
        \multirow{2}{*}{\textbf{Dataset}}
        & \multirow{2}{*}{\textbf{Method}}
        & \textbf{mAA@$10^\circ$} $\uparrow$
        & \textbf{Median error ($^\circ$)} $\downarrow$
        & \textbf{MAE ($^\circ$)} $\downarrow$ \\
        \cmidrule(lr){3-3}
        \cmidrule(lr){4-4}
        \cmidrule(lr){5-5}

        & & $e_R/e_t/e_{\max}$
          & $e_R/e_t/e_{\max}$
          & $e_R/e_t/e_{\max}$ \\
        \midrule

        \multirow{2}{*}{MegaDepth}
        & FSNet
        & 0.850 / 0.683 / 0.655 
        & 0.835 / 1.867 / 2.185
        & 5.463 / 6.615 / 8.588 \\

        & PVN
        & \textbf{0.964 / 0.850 / 0.839 }
        & \textbf{0.393 / 0.814 / 1.017 }
        & \textbf{0.787 / 4.681 / 4.243} \\
        \midrule

        \multirow{2}{*}{ScanNet++}
        & FSNet
        & 0.462 / 0.340 / 0.284 
        & 5.311 / 10.005 / 13.978 
        & 34.147 / 22.934 / 41.854 \\

        & PVN
        & \textbf{0.958 / 0.928 / 0.904}
        & \textbf{0.599 / 0.712 / 1.034}
        & \textbf{0.715 / 4.316 / 2.853} \\
        \bottomrule
    \end{tabular}
\end{table}

\begin{table}[t]
    \centering
    \caption{\textbf{Verification-guided fallback on benchmarks.}
    Each entry reports AUC at $5\degree/10\degree/20\degree$.
    We evaluate two fixed estimator orders while keeping the PVN, verification thresholds, fallback policy, and execution budget unchanged.}
    \label{tab:verifier_on_1500}
    \scriptsize
    \begin{tabular}{lcc}
        \toprule
        \textbf{Method} & \textbf{ScanNet1500} & \textbf{MegaDepth1500} \\
        \midrule
        Rank-1 
        & 0.359 / 0.581 / 0.748 
        & 0.700 / 0.814 / 0.891 \\
        
        Order 1 + PVN 
        & \textbf{0.366 / 0.592 / 0.760}
        & \textbf{0.715 / 0.828 / 0.902} \\
        
        Order 2 + PVN 
        & \textbf{0.370 / 0.596 / 0.764} 
        & \textbf{0.716 / 0.828 / 0.903} \\
        \midrule

        \rowcolor{gray!10}
        \textit{Oracle@9} 
        & 0.522 / 0.716 / 0.843 
        & 0.818 / 0.897 / 0.944 \\
        \bottomrule
    \end{tabular}
    \\[5pt]
    {\scriptsize
    ScanNet1500 orders: 1: VGGT$\to$SG-O-RS$\to$SG-I-RS$\to$RoMa-I; 2: VGGT$\to$RoMa-I$\to$SG-O-RS$\to$SG-I-RS.\\
    MegaDepth1500 orders: 1: RoMa-O$\to$SG-O-RS$\to$RoMa-I$\to$LightGlue; 2: RoMa-O$\to$RoMa-I$\to$LightGlue$\to$SG-O-RS.}
\end{table}

\noindent\textbf{Complementarity of Ranking and Verification}
% \label{sub:complementarity}
Table~\ref{tab:eval_ranker_model_selection} compares Verifier-Only selection over all nine candidates, Ranker-Only (Rank-1), Fixed-Order + PVN, and the full PoseAgent.
The fixed-order baseline sorts estimators by their average win rate across the four training datasets, weighting each dataset equally, while sharing the same PVN, thresholds, fallback policy, and $K=4$.
PoseAgent outperforms Ranker-Only and Fixed-Order + PVN across all datasets, and outperforms exhaustive Verifier-Only selection on three datasets while being only marginally lower on ScanNet++, despite executing at most four rather than all nine estimators.
These results demonstrate that ranking and verification are complementary.

\subsection{Ablation Studies}
\label{sub:ablation}
% ranker alone vs. ranker + verifier 
% best working results 
% how many iteration (top-K) 
% \subsubsection{Effect of verifier augmentation/threshold/top-k}

\paragraph{Comparison with VLM-Based Agents}
%\label{sub:eval_vlm_agent}

We evaluate VLM-based agents on ARKitScenes using Claude Sonnet-5, including a coding agent, a VLM ranker, and the VLM ranker combined with our verifier and fallback policy.
% We conduct the VLM-based agent evaluation on the ARKitScenes using Claude Sonnet-5 in three settings: a coding agent that constructs and executes a pose estimation pipeline, a VLM ranker that orders the nine candidate estimators from image pair profiles, and the VLM ranker combined with our verifier and fallback policy.
As shown in Tables~\ref{tab:comp_vlm_agent} and \ref{tab:vlm_as_ranker}, PoseAgent substantially outperforms the coding agent and achieves better ranking performance than the VLM ranker. 
% As shown in Tables~\ref{tab:comp_vlm_agent} and \ref{tab:vlm_as_ranker}, the coding agent's OpenCV-based pipeline performs substantially worse than PoseAgent.
%
% Our learned ranker also outperforms the VLM ranker in both Top-1 and Oracle@2 selection.
%
Adding our verifier improves the VLM ranking results, but PoseAgent remains superior under the same verification and fallback settings, demonstrating the complementary benefits of learned ranking and verification. 
% \input{tables/vlm_as_ranker}
% \begin{table}[H]
%     \centering
%     \caption{Relative pose estimation on ARKitScenes. Results are AUC@5\degree/10\degree/20\degree.}
%     \label{tab:comp_vlm_agent}
%     % \vspace{\baselineskip}

%     \begin{tabular}{lccc}
%         \toprule

%         % & \multicolumn{3}{c}{\textbf{AUC}} \\
%         % \cmidrule(lr){2-4}

%         \textbf{Method}
%         & \textbf{@5\degree}
%         & \textbf{@10\degree}
%         & \textbf{@20\degree} \\
%         \midrule

%         VLM Coding Agent
%         & 0.090 & 0.163 & 0.259 \\ 

%         % VLM Ranker (Top-1)
%         % & 0.369 & 0.549 & 0.691 \\ 

%         % VLM Ranker + Verifier 
%         % & \underline{0.449} & \underline{0.650} & \underline{0.795} \\

%         PoseAgent 
%         & {\bf 0.473} & {\bf 0.685} & {\bf 0.823} \\

%         % \midrule
%         % \rowcolor{lightgray}
%         % Oracle@9
%         % & 0.607 & 0.778 & 0.881 \\ 
%         \bottomrule
%     \end{tabular}
% \end{table}

\setlength{\columnsep}{10pt}
\setlength{\intextsep}{5pt}

\begin{figure}[H]
    \centering
    \includegraphics[width=0.8\linewidth]{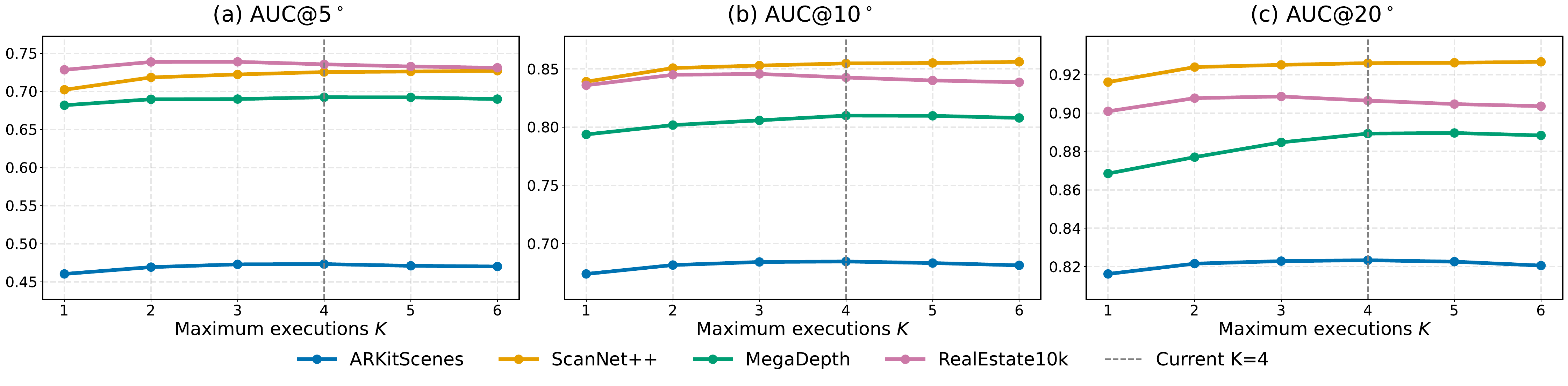}
    % \caption{The histogram records the frequency of each estimator when it gives the minimum pose error on an image pair. Seven methods are benchmarked on three evaluation datasets.}
    \caption{\textbf{Effect of the execution budget $K$ on relative pose accuracy.} We use $\tau_R=0.5^\circ$ and $\tau_t=6^\circ$ in this experiment. Dashed lines indicate the default setting with $K=4$.}
    \label{fig:ablation_topK}
\end{figure}

\begin{wraptable}{r}{0.46\textwidth}
    % \vspace{-10pt}
    \centering
    \captionsetup{font=footnotesize,skip=3pt}
    
    \caption{Results on ARKitScenes.}
    \label{tab:comp_vlm_agent}
    
    \footnotesize
    \setlength{\tabcolsep}{3pt}
    \renewcommand{\arraystretch}{0.95}
    
    % \resizebox{\linewidth}{!}{%
    \begin{tabular}{@{\hspace{10pt}}lc@{\hspace{10pt}}}
        \toprule
        \textbf{Method}
        & \textbf{AUC@5\degree/10\degree/20\degree} \\
        % & \textbf{@10\degree}
        % & \textbf{@20\degree} \\
        
        \midrule
        VLM Coding Agent
        & 0.090 / 0.163 / 0.259 \\
        
        PoseAgent
        & \textbf{0.473} / \textbf{0.685} / \textbf{0.823} \\
        \bottomrule
    \end{tabular}%
    % }
\end{wraptable}
\noindent\textbf{Effect of maximum number of estimators used}
We change the maximum number of estimators executed $K$ from 1 to 6, with ($\tau_R=0.5^\circ$) and ($\tau_t=6^\circ$). 
Figure~\ref{fig:ablation_topK} shows performance improves with additional fallback candidates, but saturates for larger $K$. 
% allowing fallback beyond the top-ranked estimator improves performance across all four datasets and all three AUC thresholds. 
% most gains come from the first few fallback candidates. 
% Further increasing ($K$) yields diminishing gains on ScanNet++. 
% It also can reduce accuracy on the other datasets, indicating that evaluating more candidates does not necessarily improve final selection. 
We therefore choose $K = 4$.% in all experiments. 
\begin{table}[!ht]
    \centering
    \begin{minipage}[c]{0.54\linewidth}
        \centering
        \caption{\textbf{Ranking and verification on ARKitScenes} (AUC@5\degree/10\degree/20\degree). Both use the same learned verifier and fallback policy with $K=4$.}
        \label{tab:vlm_as_ranker}
        \scriptsize
        \setlength{\tabcolsep}{4pt}
        \begin{tabular}{@{\hspace{10pt}}llc@{\hspace{10pt}}}
            \toprule
            \textbf{Setting} & \textbf{Method} & \textbf{@5\degree/10\degree/20\degree} \\
            \midrule
            \multirow{2}{*}{Rank-1}
                & VLM Ranker    & 0.369 / 0.549 / 0.691 \\
                & Ranker (Ours) & \textbf{0.460 / 0.674 / 0.816} \\
            \midrule
            \multirow{2}{*}{Oracle@2}
                & VLM Ranker    & 0.448 / 0.629 / 0.759 \\
                & Ranker (Ours) & \textbf{0.532 / 0.727 / 0.851} \\
            \midrule
            \multirow{2}{*}{PVN + Fallback}
                & VLM Ranker    & 0.449 / 0.650 / 0.795 \\
                & PoseAgent     & \textbf{0.473 / 0.685 / 0.823} \\
            \bottomrule
        \end{tabular}
    \end{minipage}\hfill
    \begin{minipage}[c]{0.42\linewidth}
        \centering
        \caption{\textbf{Average latency per image pair (ms).} Model initialization is excluded.}
        \label{tab:module_latency}
        \scriptsize
        \setlength{\tabcolsep}{4pt}
        \begin{tabular}{@{\hspace{10pt}}lrrr@{\hspace{10pt}}}
            \toprule
            \textbf{Dataset} & \textbf{Profiling} & \textbf{Ranking} & \textbf{Verifier} \\
            \midrule
            ARKitScenes   & 583.18 & 25.11 & 82.14 \\
            ScanNet++     & 661.42 & 25.70 & 92.09 \\
            MegaDepth     & 691.00 & 25.99 & 90.63 \\
            RealEstate10K & 613.15 & 25.28 & 84.47 \\
            \bottomrule
        \end{tabular}
    \end{minipage}
\end{table}

%\noindent\textbf{Inference Latency Analysis} for each module is given in Table~\ref{tab:module_latency}. More ablation study is provided in Appendix~\ref{app:more_ablation}.

\noindent\textbf{Limitations and Future Work} PoseAgent increases overall latency (Table~\ref{tab:module_latency}) as it may run multiple estimators, especially for challenging image pairs.
The profiling stage can be further optimized for efficiency.
%PVN may overestimate or underestimate pose errors, leading to incorrect verification decisions, as shown in Fig.~\ref{fig:exp4_viz}.
Future work includes faster profiling, extensions to absolute camera pose estimation, incorporating more pose estimators, and developing better fallback policies.
% PoseAgent may execute multiple estimators, especially for difficult image pairs, which increases inference cost.
% PVN can also overestimate or underestimate pose errors, which may cause good candidates to be rejected or poor candidates to be accepted, as shown in Fig.~\ref{fig:exp4_viz}.
% \input{tables/latency}

% \input{sections/6experiments_backup}
% \input{sections/7results}
\section{Conclusion}

In this work, we introduce \textbf{PoseAgent}, an agentic framework for relative camera pose estimation that dynamically orchestrates specialized pose estimators through learnable \textit{ranking} and \textit{verification}.
Given an arbitrary image pair, a profiling agent first extracts semantic and geometric characteristics that capture factors relevant to pose estimation.
A learned ranking agent then predicts the relative competence of multiple pose estimators conditioned on the resulting image-pair profile.
The top-ranked estimator is executed, and its predicted pose is assessed by a learned verification agent without access to ground-truth camera pose.
When verification fails, PoseAgent adaptively invokes lower-ranked estimators until a reliable pose estimate is obtained.
% outperform VLM agent
Our verification network substantially outperforms previous methods for pose verification.
Extensive experiments on ARKitScenes, MegaDepth, ScanNet++, and RealEstate10K demonstrate that PoseAgent consistently outperforms individual standalone pose estimators.
Moreover, PoseAgent outperforms VLM-based agents for relative camera pose estimation, demonstrating the effectiveness of ranking and verification.

\clearpage

\subsection*{Reproducibility Statement}
We provide comprehensive experimental and implementation details to facilitate reproducibility. Section 4 describes datasets, PVN optimization settings, ablation studies, and comparison with VLM-based agents. The appendices specify implementation details including training settings, profiling configuration, system prompts, and example visualization. The results distinguish development evaluations from independent confirmation. The accompanying NumPy package reproduces the new offline curves and paired intervals from sufficient anonymous statistics. 

\subsection*{AI use statement}

% \todo{add statement} 
%(This section is \textbf{required} and does not count toward the page limit.)
% AI coding assistants were used to assist with language editing of author-written text, code drafting and debugging for the agentic system, as well as figure design creation. All AI-assisted text, code, and figures were reviewed and revised by the authors, who take full responsibility for the content of this paper.

In this work, we used generative AI tools to formulate mathematical claims, assist in translation and qualitative and thematic data analysis. 
We have not used generative AI tools for research idea formulation and dataset generation and preprocessing is not applicable to this work. 
Additionally, we used generative AI tools for editing a research paper to improve readability and formatting references. 
We have reviewed all AI-assisted work. For example, we refer AI-generated figure style and make figures on our own. 
We also use it to assist in code debugging to make sure the implementation runs successfully.
We take responsibility for the final content of this work, including text, claims, or artifacts produced with the aid of generative AI.

\bibliographystyle{iclr2027_conference}
\bibliography{iclr2027_conference}

\appendix
\clearpage
\newtcolorbox{promptbox}[1]{
  enhanced,
  breakable,
  title={#1},
  colback=white,
  colframe=black!75,
  colbacktitle=black!80,
  coltitle=white,
  fonttitle=\bfseries\large,
  fontupper=\normalsize,
  boxrule=0.8pt,
  arc=2.5mm,
  left=4mm,
  right=4mm,
  top=3mm,
  bottom=3mm,
  lefttitle=4mm,
  righttitle=4mm,
  toptitle=2mm,
  bottomtitle=2mm
}

\section{Appendix}
% add some quantitative results 
% visualize different images preferred by various methods 
\subsection{Implementation Details}
\label{app:implementation_details}
We train one ranker and one PVN using data pooled from all four datasets.
%
% We jointly train a single ranker and a single PVN across all four datasets.
Validation pairs are randomly held out from the training scenes, while the test scenes are scene-disjoint from both the training and validation sets.
Dataset statistics are summarized in Table~\ref{tab:datasets}.
\setlength{\intextsep}{0.5\baselineskip}
\begin{table}[H]
    \centering
    \caption{\textbf{Datasets for PoseAgent.} It uses four datasets covering indoor and outdoor environments.}
    \label{tab:datasets}
    \footnotesize
    \begin{tabular}{cccc}
        \toprule
        \textbf{Datasets} & \textbf{Scene Type} & \textbf{Training Pairs} & \textbf{Test Pairs} \\
        \midrule
        
         ARKitScenes  & Indoor & 20,000 & 1,095 \\
         ScanNet++    & Indoor & 18,202 & 1,798 \\ 
         MegaDepth    & Outdoor & 17,879 & 2,121 \\
         RealEstate10K  & Indoor \& Outdoor & 20,000 & 5,449 \\
         \bottomrule
    \end{tabular}
\end{table}

Before verification and evaluation, their predictions are converted to the same camera convention, from camera 1 to camera 2.
% Validation pairs are randomly held out, which test scenes are disjoint from both training and validation scenes. 
% PVN checkpoints are selected by the lowest validation loss. 
% PoseAgent operates over nine candidate relative pose estimators, including SuperGlue-Indoor with image resizing (SG-I-RS)~\citep{sarlin2020superglue}, SuperGlue-Outdoor with image resizing (SG-O-RS), SuperGlue-Indoor (SG-I), SuperGlue-Outdoor (SG-O), LightGlue~\citep{lindenberger2023lightglue}, RoMa-Indoor(RoMa-I)~\citep{edstedt2024roma}, RoMa-Outdoor (RoMa-O), Reloc3r~\citep{dong2025reloc3r}, and VGGT~\citep{wang2025vggt}. 
% For the resized SuperGlue variants, the longer image dimension is resized to 1600 pixels. 
% We use the pretrained weights for all candidate estimators and follow their inference configurations. 
% All predicted relative poses are converted to the same pose convention, from camera 1 to camera 2, before being passed to the ranker and verifier. 

We implement the Ranking Agent using XGBRanker with the \texttt{rank:ndcg} objective, 300 trees, a maximum depth of 6, and a learning rate of 0.05.
%
% The model contains 300 trees with a maximum depth of 6 and is trained using a learning rate of 0.05.
For the PVN, we freeze the pretrained Reloc3r encoder and decoder and train only the newly introduced pose-conditioned modules and pose error prediction heads.
% prediction heads for rotation and translation errors. 
Training runs for 60 epochs on 7 NVIDIA H100 GPUs using AdamW, with a batch size of 32 per GPU, an initial learning rate of $3\times 10^{-4}$, a weight decay of 0.1, and cosine learning-rate scheduler.

% To increase the diversity of pose candidates, we employ two augmentation strategies. With the probability $p=0.35$, we perturb the original candidate pose. With the probability $q=0.15$, we synthesize a candidate pose around the ground truth pose by independently sampling its rotation and translation-direction. The remaining 50\% of candidates are used without augmentation.  

\subsection{Profiling tools and extracted features}
\label{app:profiling_tools}

Table~\ref{tab:profiling_tools} summarizes the computer vision tools used in the Image Pair Profiling Agent. The resulting profile describes each image pair's appearance and scene features, as well as cross-view correspondence and geometry features. 
We also derive pair-level statistics, such as differences between two images and heuristic difficulty and degeneracy labels. 
\setlength{\intextsep}{0.5\baselineskip}
\begin{table}[H]
    \centering
    \caption{Computer vision tools used to construct image-pair profiles.}
    \label{tab:profiling_tools}
    \small
    \begin{tabular}{p{0.19\linewidth}p{0.75\linewidth}}
        \toprule
        \textbf{Tool} & \textbf{Extracted information} \\
        \midrule
        
        % OpenCV - Laplacian, FAST, ORB, RANSAC 
        % & Image resolution, brightness, contrast, blur, entropy, and
        % keypoint density; ORB matches and their spatial spread; homography,
        % fundamental-matrix, and essential-matrix inlier ratios; scale change
        % and motion cues. \\
        OpenCV
        & Image quality and texture (Laplacian, FAST); 
        ORB matches and RANSAC geometric inliers; 
        Farneback optical flow variation. \\ 
        % & Laplacian, FAST, and ORB features; RANSAC inlier ratios \\ 

        % OpenCV
        % & Image quality, texture, correspondence geometry and motion variation \\ 
        
        CLIP 
        & Zero-shot scene category, scene type \\
        
        DINOv2 
        & Appearance change measured from the cosine distance between
        image embeddings. \\
        
        SegFormer 
        & Semantic category pixel ratios, scene composition \\
        
        LoFTR 
        & Spatial coverage of confident matches, visual overlap \\
        
        SuperPoint 
        & Keypoints and descriptor matching statistics, texture and match quality \\
        \bottomrule
    \end{tabular}
\end{table}

\subsection{Ranker}
\label{app:ranking_curves}

\begin{figure}[h]
    \centering
    \includegraphics[width=\linewidth]{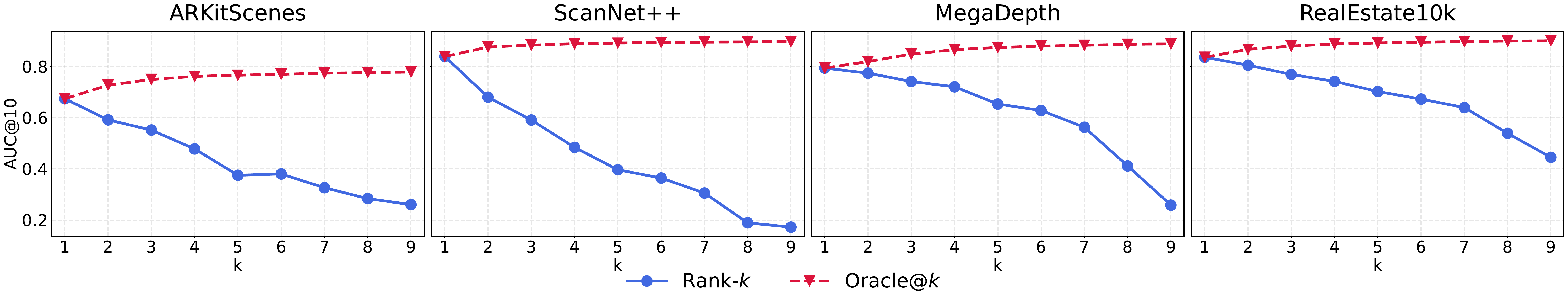}
    % \caption{The histogram records the frequency of each estimator when it gives the minimum pose error on an image pair. Seven methods are benchmarked on three evaluation datasets.}
    \caption{\textbf{Performance across predicted estimator rank and oracle top-$k$ selection.} Rank-$k$ reports the AUC@10\degree using the estimator at predicted rank $k$; Oracle@$k$ selects the best prediction among the top-$k$ ranked estimators.}
    \label{fig:rank_oracle_auc10}
\end{figure}

Figure~\ref{fig:rank_oracle_auc10} shows that Rank-$k$ performance generally decreases with rank, while Oracle@$k$ improves as more candidates are considered. 
This indicates that our ranker produces a meaningful ordering of candidate estimators. 
Oracle@9 provides upper bound, but requires evaluating all candidate estimators. This motivates the verifier introduced in the section~\ref{sub:eval_verifier}, which aims to identify unreliable Rank-1 predictions and selectively fall back to lower-ranked candidates only when necessary.

\subsection{VLM as a coding agent}
We use Claude Sonnet-5 with maximum output of 2,048 tokens, as a VLM coding agent baseline for relative camera pose estimation.
For each image pair, we provide two images and their camera intrinsics. 
The model was given access to a code-execution tool and instructed to detect and match image features and return a relative rotation matrix and translation vector. 
The requested output is a JSON object containing $\mathbf{R}$ and $\mathbf{t}$.

The prompt for code execution and output formatting is shown below. 

\begin{promptbox}{VLM as a Coding Agent}

    Your role is a research assistant specializing in computer vision.
    
    \medskip
    Use available approaches to:
    \begin{enumerate}[leftmargin=*, itemsep=1pt, topsep=3pt]
        \item Detect and match feature points between the two images.
        \item Estimate the essential matrix using the matched points and
              the provided camera intrinsics.
        \item Decompose the essential matrix to obtain the relative rotation
              matrix (R) and translation vector (t).
        \item If failed, return a JSON object with R as an identity matrix
              and t as a zero vector.
    \end{enumerate}
    
    \medskip
    \textbf{CRITICAL INSTRUCTIONS:}
    \begin{itemize}[leftmargin=*, itemsep=1pt, topsep=3pt]
        \item Code NEVER starts with
              \texttt{\detokenize{`py\n}}
              or ends with \texttt{\detokenize{`}}.
        \item Code starts with: \texttt{import} \ldots
        \item NEVER wrap code in triple backticks
              (\texttt{\detokenize{```}}) or use markdown.
        \item You MUST write and execute Python code using the code interpreter.
        \item After execution, you MUST output ONLY a valid JSON object
              in the exact format below.
        \item DO NOT include any other text, explanation, code, or markdown
              before or after the JSON.
        \item The entire response must be parseable as JSON.
        \item When generating code for the \texttt{code\_interpreter},
              output ONLY raw Python code.
        \item NEVER include comments like \texttt{\detokenize{"# Step 1"}}
              unless necessary.
        \item The code must be immediately executable.
        \item Note: \texttt{cv2.decomposeEssentialMat} only has 3 returns
              instead of 4 or 2.
        \item (STRICT) If you use \texttt{cv2.findEssentialMat}, you can use
              \texttt{cv2.LORANSAC} as the method.
    \end{itemize}
    
    \medskip
    The exact image filenames available in the code interpreter are:
    
    Image A: \texttt{\{image1\_name\}}\\
    Image B: \texttt{\{image2\_name\}}
    
    You MUST use these exact filenames, including all file extensions,
    when calling \texttt{cv2.imread()}.
    
    \medskip
    Task: Estimate the relative camera pose between the two attached
    images (Image A and Image B) using the python tool.
    
    \medskip
    Camera Intrinsics:
    
    Image 1: fx=\texttt{\{fx1\}}, fy=\texttt{\{fy1\}},
    cx=\texttt{\{cx1\}}, cy=\texttt{\{cy1\}}\\
    Image 2: fx=\texttt{\{fx2\}}, fy=\texttt{\{fy2\}},
    cx=\texttt{\{cx2\}}, cy=\texttt{\{cy2\}}
    
    \medskip
    Output Format (STRICT): Provide the final result ONLY as a single
    JSON object for rotation matrix (R) and translation vector (t).
    
    Output Format results must attach with a tag [JSON schema] at the
    beginning. For example:
    
    \medskip
    {\ttfamily\footnotesize
    [JSON RESULTS]:\{\{"R": [[r11, r12, r13], [r21, r22, r23],
    [r31, r32, r33]], "t": [[t1], [t2], [t3]]\}\}}
\end{promptbox}

\subsection{VLM as a ranker}
We evaluate a VLM ranking baseline. 
For each image pair, the model receives both images and a JSON object containing the extracted profiling features. 
A system prompt describes the nine candidate pose estimators and the meaning of the profiling features. 
The model is instructed to return a JSON array containing all nine candidates, ordered from the most to least suitable. 

We use Claude Sonnet-5 and set the maximum output to 512 tokens. 
The complete system prompt and each pair's content prompt are shown below. 

\begin{promptbox}{System Prompt}
    You need to rank relative camera pose estimators for each image pair.\\
    \par\medskip
    Goal: Use provided two images and their extracted profiling features,\\
    return all 9 candidates from the most suitable pose estimator (rank1) to the least suitable one.\\
    \par\medskip
    Explanation of 9 candidate pose estimators:\\
    \textbullet\ superglue\_indoor / superglue\_outdoor: SuperPoint + corresponding indoor or outdoor SuperGlue weights, followed by LO-RANSAC.\\
    \textbullet\ superglue\_indoor\_resize1600 / superglue\_outdoor\_resize1600: the same model and weights as superglue\_indoor/superglue\_outdoor,\\
        but resize the longest dimension to 1600.\\
    \textbullet\ lightglue: LightGlue model, followed by LO-RANSAC.\\
    \textbullet\ roma\_indoor / roma\_outdoor: dense RoMa correspondence with indoor / outdoor weights, followed by LO-RANSAC.\\
    \textbullet\ reloc3r: end-to-end relative-pose regression.\\
    \textbullet\ vggt: 3D foundation model that also can predict relative-camera pose.\\
    \par\medskip
    \textbf{FEATURES:}\\
    \textbf{Quality:} blur\_score is mean Laplacian variance (higher normally means sharper,\\
    despite the name); brightness\_mean/std describe intensity; avg\_contrast and\\
    img1/2\_contrast are std/mean; avg\_entropy is intensity richness; *\_diff and\\
    texture\_imbalance measure cross-view mismatch; keypoint\_count and\\
    avg\_kp\_uniformity describe amount and spatial spread of detected texture.\\
    \par\medskip
    \textbf{Correspondence:} visual\_overlap is spatial coverage of matches and, in accurate\\
    mode, normally equals visual\_overlap\_learned. The learned value is the occupied\\
    fraction of an 8x8 grid from confident LoFTR matches. appearance\_change normally\\
    equals appearance\_change\_learned in accurate mode; the learned value is DINOv2\\
    cosine distance (0 similar, larger more different). Treat each duplicated pair\\
    as one signal. num\_putative\_matches is ORB cross-check count. match\_quality is\\
    that count times F-inlier ratio. sp\_num\_matches counts SuperPoint matches passing\\
    Lowe ratio \textless{}0.8; lower sp\_lowe\_ratio\_mean and higher sp\_match\_confidence\_mean,\\
    sp\_confident\_ratio, or sp\_match\_spatial\_consistency usually indicate more\\
    distinctive/consistent matches. Standard deviations measure variability.\\
    appearance\_per\_overlap=log(1+appearance\_change/(visual\_overlap+0.001)).\\
    \par\medskip
    \textbf{Geometry:} h\_inlier\_ratio and e\_inlier\_ratio are homography and essential-matrix\\
    RANSAC inlier fractions. h\_f\_ratio=H/F inlier ratio; planarity\_score=H-F;\\
    geometry\_quality=max(H,F). rotation\_angle is recovered rotation in degrees.\\
    scale\_change\_ratio is median matched scale image2/image1 and\\
    scale\_change\_magnitude=abs(log(ratio)). baseline\_metric/estimate are heuristic:\\
    they may come from triangulated inverse depth or pixel displacement and are not\\
    metric translation.\\
    \par\medskip
    \textbf{Motion/difficulty:} motion\_h\_score/f\_score repeat H/F evidence.\\
    motion\_rotation\_component and motion\_translation\_component are homography-based\\
    heuristics, not true motion magnitudes. flow\_spatial\_var is variance of mean\\
    Farneback-flow magnitude over four quadrants. motion\_dominant, difficulty,\\
    degeneracy, motion\_is\_degenerate, and degeneracy\_confidence are profiler\\
    summaries; corroborate them with images and underlying measurements.\\
    \par\medskip
    \textbf{Scene:} img1/2\_scene\_type are CLIP categories. building/wall/sky/vegetation ratios\\
    are mean SegFormer pixel fractions. lap\_var\_min\_mean summarizes the weakest-\\
    texture quadrant across the two images. Semantic predictions can be wrong.\\
    \par\medskip
    \textbf{Output Format:} Return JSON object containing the candidate model names ordered from\\
    the most suitable pose estimator to the least one.\\
    \par\medskip
    Use each of the following model names exactly once:\\
    \textbullet\ superglue\_outdoor\_resize1600\\
    \textbullet\ superglue\_indoor\_resize1600\\
    \textbullet\ superglue\_outdoor\\
    \textbullet\ reloc3r\\
    \textbullet\ roma\_indoor\\
    \textbullet\ lightglue\\
    \textbullet\ roma\_outdoor\\
    \textbullet\ superglue\_indoor\\
    \textbullet\ vggt\\
    \par\medskip
    Return exactly:\\
    \{"ranking": ["best\_model\_name","second\_best\_model\_name","...","worst\_model\_name"]\}\\
    
    Do not rename, abbreviate, omit, or repeat any model.\\
    \end{promptbox}
    
    \begin{promptbox}{User Message Template}
    [\textcolor{blue}{Image A attached}]\par
    [\textcolor{blue}{Image B attached}]\par
    Rank this pair. Input JSON:\\
    \texttt{\{profiling\_features\_as\_compact\_JSON\}}
\end{promptbox}

\subsection{Example Visualization}
\label{app:visualization}
We show the execution behavior of PoseAgent on selected test pairs. 
For each image pair, we show the predicted estimator ordering, the candidates actually executed, the pose errors predicted by PVN, and the resulting acceptance or fallback decisions. 

% \begin{figure}[h]
%     \centering
%     \includegraphics[page=1, width=\linewidth]{figures/example_visualization.pdf}
%     \caption{\textbf{Examples of our framework.} Given an image pair (top), the Profiling Agent constructs features to rank candidate pose estimators. PVN then evaluates the ranked pose hypotheses using predicted errors. In this example, the third-ranked candidate, Reloc3r, is selected. Ground-truth errors are shown for evaluation only.}
%     \label{fig:exp1_viz}
% \end{figure}

\begin{figure}[H]
    \centering
    \includegraphics[page=1,width=0.99\linewidth]{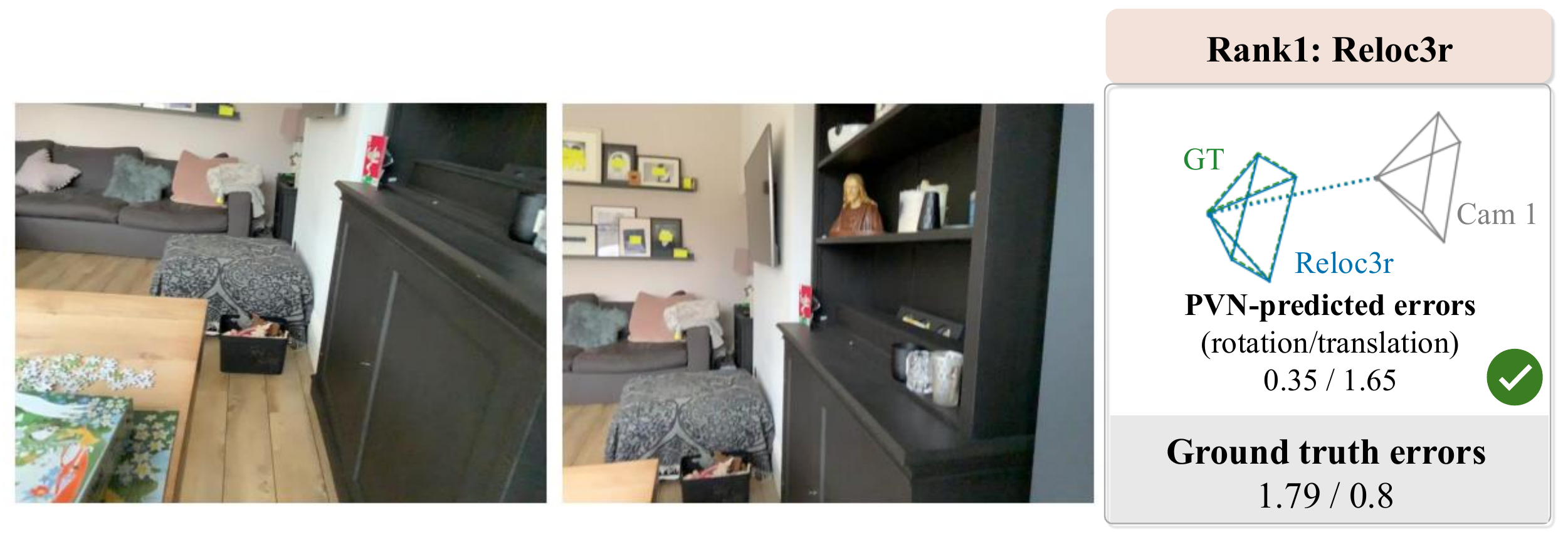}
    \caption{\textbf{Successful early exit.} PoseAgent ranks Reloc3r first and PVN accepts its estimate. The execution terminates after a single model. This example shows how accurate ranking combined with the verification avoids unnecessary model executions.}
    \label{fig:exp1_viz}
\end{figure}

\begin{figure}[H]
    \centering
    \includegraphics[page=2,width=\linewidth]{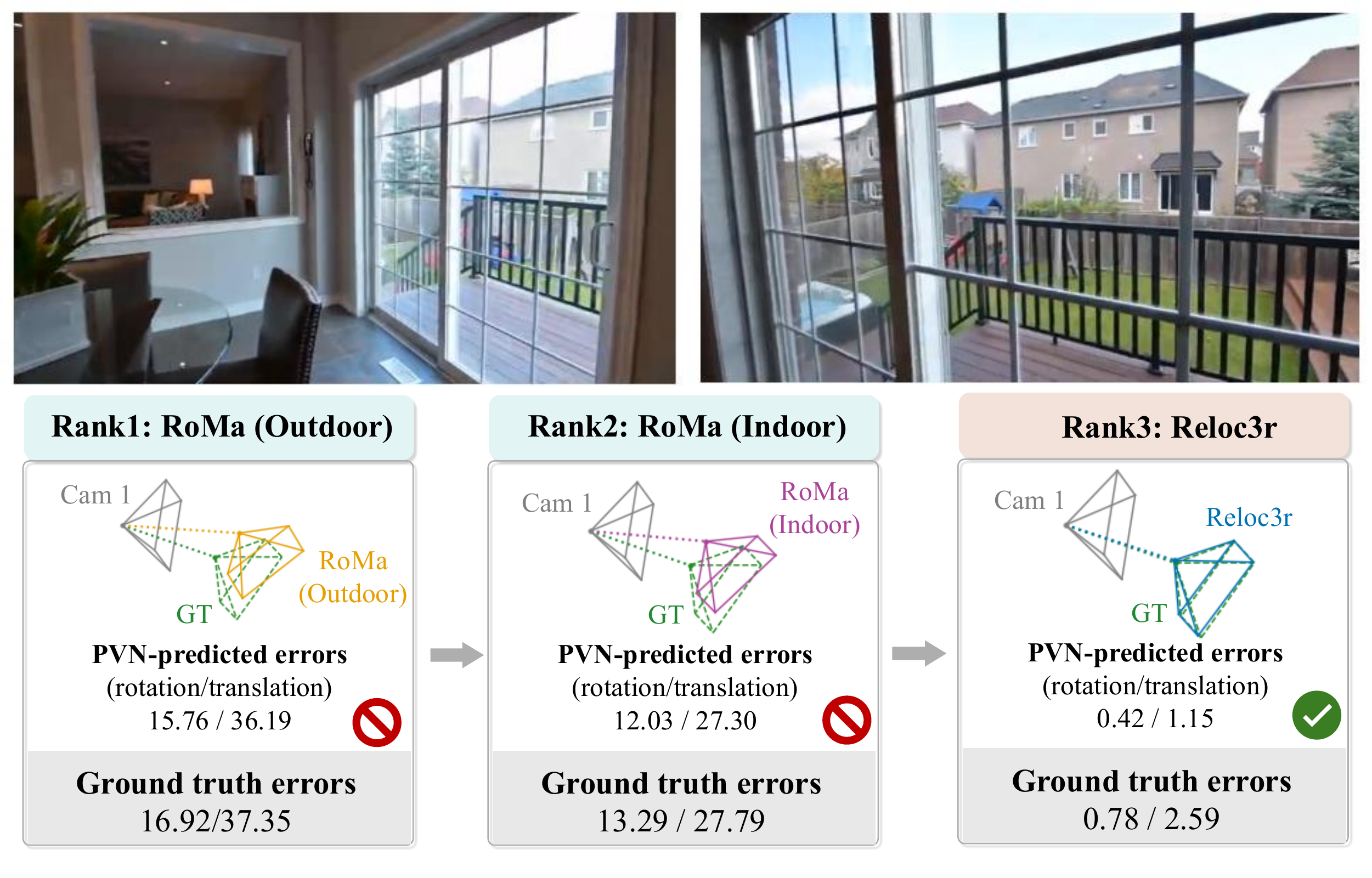}
    \caption{\textbf{Recovery from inaccurate top-ranked candidates.} PVN rejects the first two candidates, RoMa (Outdoor) and RoMa (Indoor). It then accepts Rank-3 Reloc3r. PoseAgent reduces the final error from the Rank-1 error of $37.35\degree$ to $2.59\degree$ after three model executions.}
    \label{fig:exp2_viz}
\end{figure}

\begin{figure}[H]
    \centering
    \includegraphics[page=3,width=\linewidth]{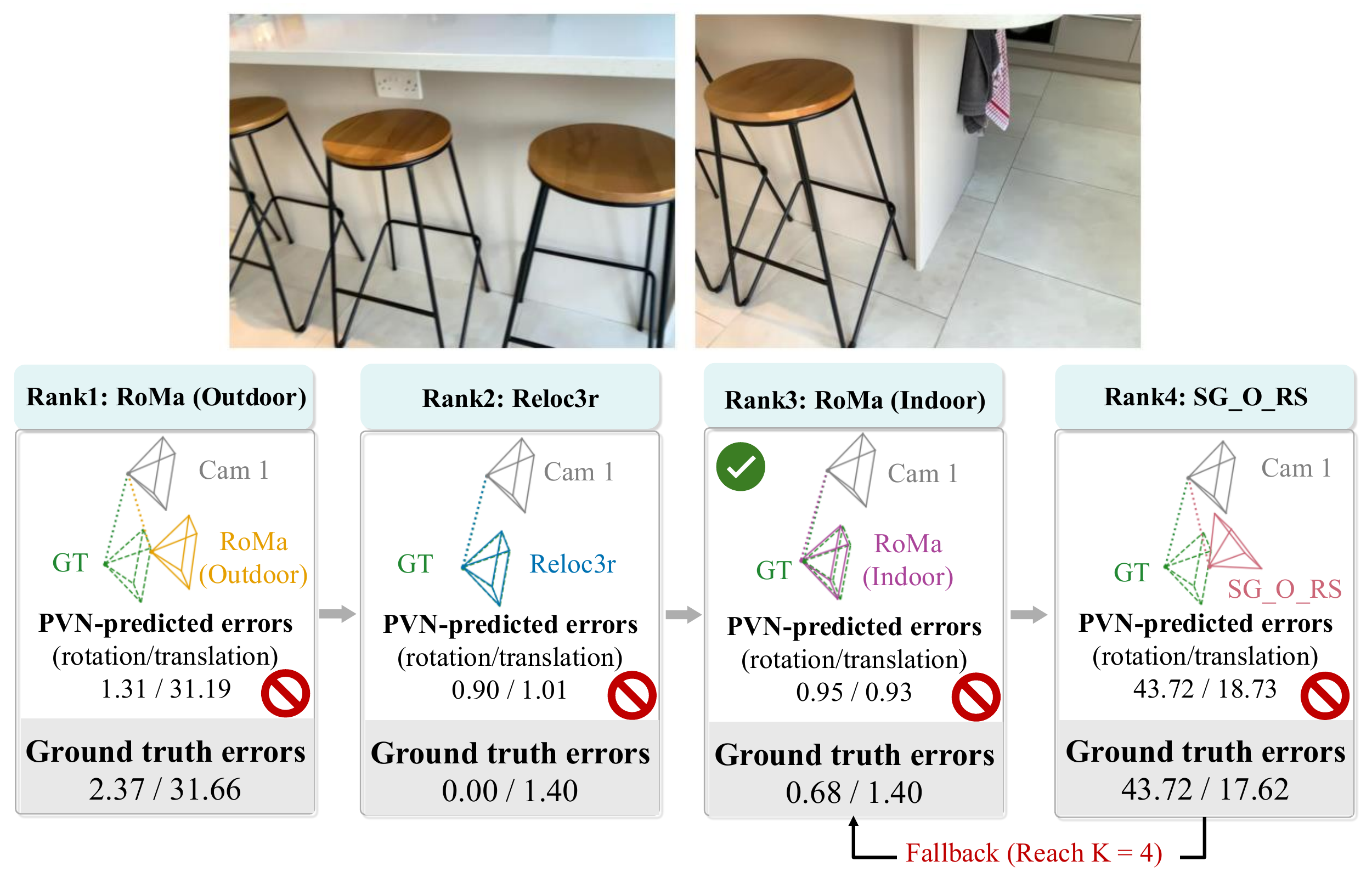}
    \caption{\textbf{Recovery through fallback.} None of the Top-4 candidate models satisfies the strict PVN acceptance thresholds, so PoseAgent executes all four models before invoking fallback. Based on the predicted errors, fallback policy selects the earlier Rank-3 RoMa (Indoor) estimate, reducing the actual error from the Rank-1 error of $31.66\degree$ to $1.40\degree$. This example shows how fallback can recover a useful pose even when no candidate is directly accepted.}
    \label{fig:exp3_viz}
\end{figure}

\begin{figure}[H]
    \centering
    \includegraphics[page=4,width=\linewidth]{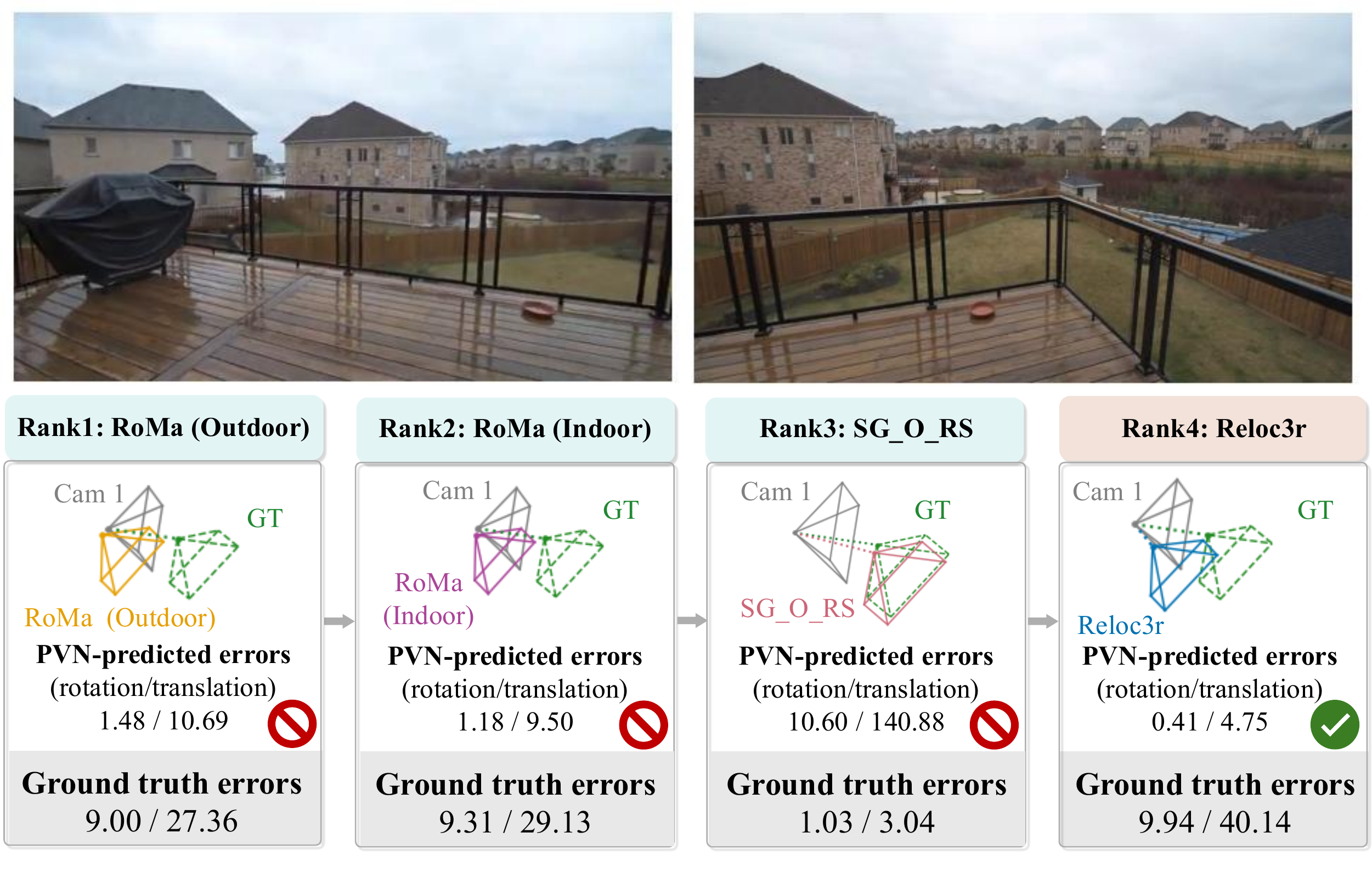}
    \caption{\textbf{Failure caused by verifier prediction accuracy.} PVN rejects the Rank-3 model's estimate despite its low actual error. It subsequently accepts Rank-4 Reloc3r. This failure example indicates that PVN underestimates the pose error on good candidates, but overestimate on poor ones.}
    \label{fig:exp4_viz}
\end{figure}

\clearpage

\subsection{More ablation studies}

\label{app:more_ablation}

\noindent \textbf{Effect of profiling features}
We examine how different profiling features affect estimator ranking. \textit{Scene features} use the predicted scene types of two input images. 
\textit{Geometry features} correspondence, overlap, keypoint distribution, motion, and geometric consistency features. 
\textit{All features} use all profiling features. 
We keep the same configuration for each ranker, PVN, verification thresholds, and the fallback policy as well as the maximum number of estimators ($K$=4).

As shown in Table~\ref{tab:profiling_ablation}, \textit{Geometry features} outperforms \textit{Scene features} across all four datasets on AUC@5/10/20\degree. 
Although coarse scene type information alone already achieves good results, they are still insufficient to account for the performance of PoseAgent. 
\textit{Geometry features} matches \textit{All features} on ARKitScenes and ScanNet++, while \textit{All features} result provides modest, consistent improvements on the other datasets. 
These results suggest that geometric profiling features further supply additional information needed for effective estimator ranking.

\begin{table}[t]
    \centering
    \caption{
        \textbf{Ablation of profiling features}.
        Each ranker is trained using the specified feature subset,
        while PVN, verification thresholds,
        the fallback policy, and the maximum execution budget
        ($K=4$) are kept unchanged.
        Best results are shown in bold,
        including ties at the reported precision.
    }
    \label{tab:profiling_ablation}
    \small
    % \setlength{\tabcolsep}{5pt}
    % \vspace{\baselineskip}
    \begin{tabular}{llccc}
        \toprule
        \textbf{Dataset} & \textbf{Profiling configuration}
        & \textbf{AUC@5\degree} & \textbf{AUC@10\degree} & \textbf{AUC@20\degree} \\
        \midrule

        \multirow{3}{*}{ARKitScenes}
        & Scene Features
        & 0.467 & 0.680 & 0.821 \\
        & Geometry Features
        & \textbf{0.473} & \textbf{0.685} & \textbf{0.824} \\
        & All features
        & \textbf{0.473} & \textbf{0.685} & 0.823 \\
        \midrule

        \multirow{3}{*}{ScanNet++}
        & Scene Features
        & 0.694 & 0.838 & 0.917 \\
        & Geometry Features
        & \textbf{0.726} & \textbf{0.855} & \textbf{0.926} \\
        & All features
        & \textbf{0.726} & \textbf{0.855} & \textbf{0.926} \\
        \midrule

        \multirow{3}{*}{MegaDepth}
        & Scene Features
        & 0.682 & 0.803 & 0.886 \\
        & Geometry Features
        & 0.691 & 0.808 & 0.888 \\
        & All features
        & \textbf{0.693} & \textbf{0.810} & \textbf{0.889} \\
        \midrule

        \multirow{3}{*}{RealEstate10K}
        & Scene Features
        & 0.673 & 0.807 & 0.887 \\
        & Geometry Features
        & 0.733 & 0.840 & 0.904 \\
        & All features
        & \textbf{0.736} & \textbf{0.843} & \textbf{0.907} \\
        \bottomrule
    \end{tabular}
\end{table}

\end{document}